\pdfoutput=1

\documentclass[11pt]{article}

\usepackage{acl}

\usepackage{times}
\usepackage{latexsym}
\usepackage{arydshln}
\usepackage{graphicx}
\usepackage{subcaption}
\usepackage{hyperref}
\usepackage{booktabs,arydshln}
\usepackage{amsmath}
\usepackage{enumitem}
\usepackage{multirow}
\usepackage[export]{adjustbox}
\usepackage[utf8]{inputenc}
\usepackage{soul}
\usepackage{cleveref}
\usepackage{xurl}

\Crefname{figure}{Fig}{Figs}
\Crefname{appendix}{App}{Apps}
\Crefname{table}{Tab}{Tabs}
\Crefname{section}{Sec}{Secs}
\Crefname{equation}{Eq}{Eqs}

\usepackage[T1]{fontenc}

\usepackage[utf8]{inputenc}

\usepackage{microtype}

\newcommand{\caribbeandataset}{Caribbean newspapers }

\newcommand\nnfootnote[1]{%
  \begin{NoHyper}
  \renewcommand\thefootnote{}\footnote{#1}%
  \addtocounter{footnote}{-1}%
  \end{NoHyper}
}

\title{Measuring Intersectional Biases in Historical Documents \\
\normalsize \textcolor{Bittersweet!60}{WARNING: This paper shows dataset samples that are racist in nature}}

\author{
Nadav Borenstein$^{\ast 1}$ \quad
Karolina Stańczak$^{\ast 1}$ \quad
Thea Rolskov$^{2}$ \quad
Natália da Silva Perez$^{3}$ \\
\textbf{Natacha Klein Käfer}$^{1}$ \quad
\textbf{Isabelle Augenstein}$^{1}$ \\
$^1$University of Copenhagen \quad
$^2$Aarhus University \quad
$^3$Erasmus University Rotterdam \\
{\tt \href{mailto:nadav.borenstein@di.ku.dk}{nadav.borenstein@di.ku.dk}} \quad
{\tt \href{mailto:ks@di.ku.dk}{ks@di.ku.dk}} \quad 
{\tt \href{mailto:201706833@post.au.dk}{201706833@post.au.dk}} \\ 
{\tt \href{mailto:dasilvaperez@eshcc.eur.nl}{dasilvaperez@eshcc.eur.nl}} \quad
{\tt \href{mailto:nkk@teol.ku.dk}{nkk@teol.ku.dk}} \quad
{\tt \href{mailto:augenstein@di.ku.dk}{augenstein@di.ku.dk}} \quad 
}

\begin{document}
\maketitle

\begin{abstract}
Data-driven analyses of biases in historical texts can help illuminate the origin and development of biases prevailing in modern society.
 However, digitised historical documents pose a challenge for NLP practitioners as these corpora suffer from errors introduced by optical character recognition (OCR) and are written in an archaic language.
In this paper, we investigate the continuities and transformations of bias in historical newspapers published in the Caribbean during the colonial era (18th to 19th centuries). Our analyses are performed along the axes of gender, race, and their intersection. We examine these biases by conducting a temporal study in which we measure the development of lexical associations using distributional semantics models and word embeddings. Further, we evaluate the effectiveness of techniques designed to process OCR-generated data and assess their stability when trained on and applied to the noisy historical newspapers.
We find that there is a trade-off between the stability of the word embeddings and their compatibility with the historical dataset. 
We provide evidence that gender and racial biases are interdependent, and their intersection triggers distinct effects. These findings align with the theory of intersectionality, which stresses that biases affecting people with multiple marginalised identities compound to more than the sum of their constituents. \nnfootnote{* Equal contribution.}

\vspace{1.5em}
\hspace{.5em}\includegraphics[width=1.25em,height=1.25em]{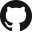}\hspace{.75em}\parbox{\dimexpr\linewidth-2\fboxsep-2\fboxrule}{\url{https://github.com/copenlu/intersectional-bias-pbw}}
\vspace{-.5em}
\end{abstract}

\section{Introduction}
\label{sec:introduction}
\begin{figure}[ht]
    \centering

        \includegraphics[width=\columnwidth, trim={0.25cm 0 0 0},clip]{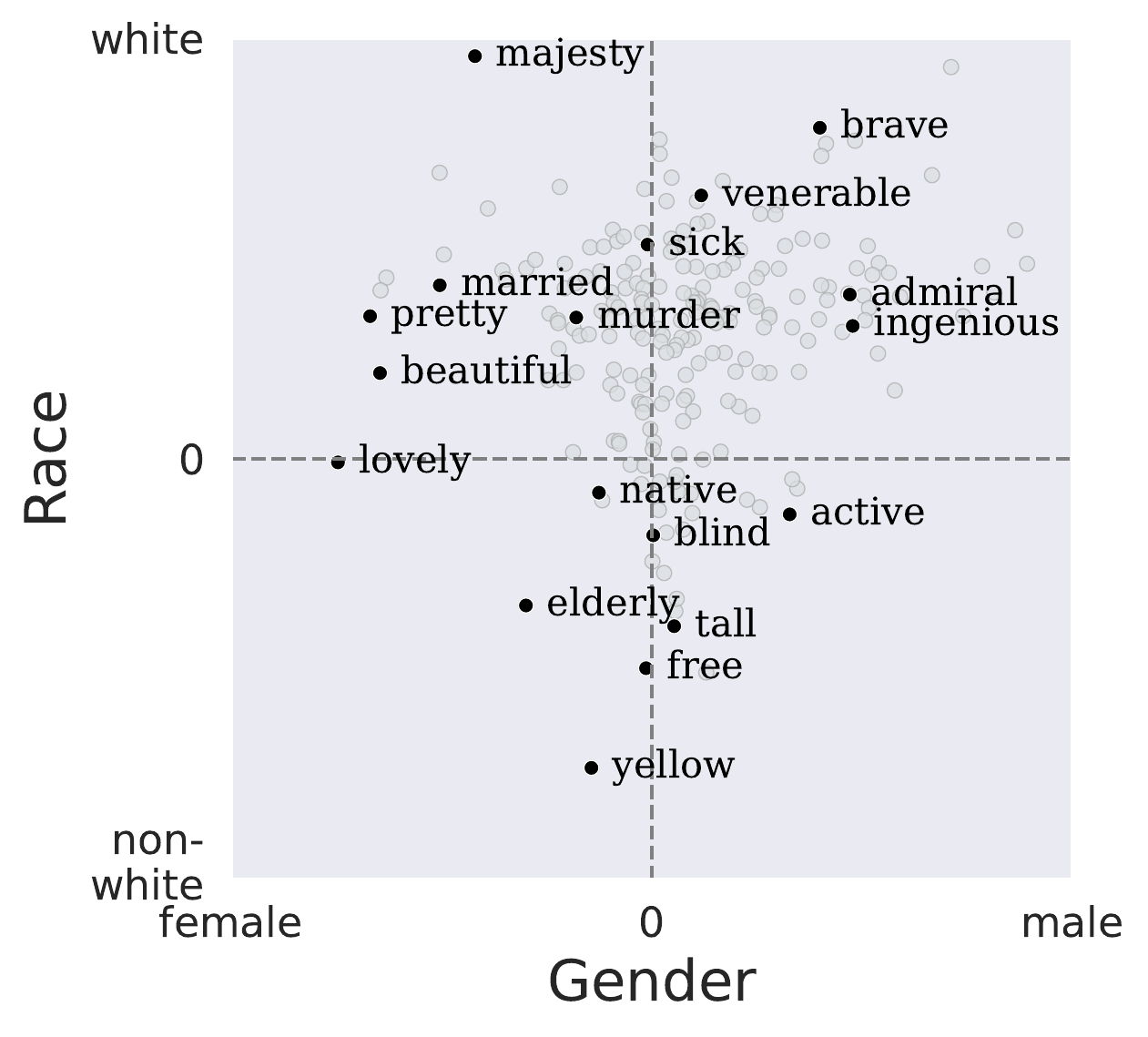}
         \caption{PMI analysis of our historical corpora. Words are placed on the intersectional gender/race plane.}
         \label{fig:general_bias_descriptors}
         
\end{figure}

The availability of large-scale digitised archives and modern NLP tools has enabled a number of sociological studies of historical trends and cultures \citep{garg2018stereotypes, Kozlowski_2019,michel2011quantitative}. 
Analyses of historical biases and stereotypes, in particular, can shed light on past societal dynamics and circumstances \citep{sullam2022representation} and link them to contemporary challenges and biases prevalent in modern societies \citep{payne2019slavery}. 
For instance, \citet{payne2019slavery} consider implicit bias as the cognitive residue of past and present structural inequalities and highlight the critical role of history in shaping modern forms of prejudice.

Thus far, previous research on bias in historical documents focused either on gender \citep{rios-etal-2020-quantifying, wevers-2019-using} or ethnic biases \citep{sullam2022representation}. While \citet{garg2018stereotypes} separately analyse both, 
their work does not engage with their intersection.
Yet, in the words of \citet{crenshaw_mapping_1995}, intersectional perspective is important because  ``the intersection of racism and sexism factors into black women’s lives in ways that cannot be captured wholly by looking separately at the race or gender dimensions of those experiences.''

Analysing historical documents poses particular challenges for modern NLP tools \citep{nadav2023, ehrmann-etal-2020-language}. Misspelt words due to wrongly recognised characters in the digitisation process, 
and archaic language unknown to modern NLP models, i.e.
historical variant spellings and words that became obsolete in the current language, increase the task's complexity \citep{bollmann-2019-large, linharespontes:hal-02557116,piotrowski2012natural}. However, while most previous work on historical NLP acknowledges the unique nature of the task, only a few address them within their experimental setup. 


In this paper, we address the shortcomings of previous work and make the following contributions: 
(1) To the best of our knowledge, this paper presents the first study of historical language associated with entities at the intersections of two axes of oppression: race and gender. We study biases associated with identified entities on a word level, and to this end, employ distributional models and analyse semantics extracted from word embeddings trained on our historical corpora.
(2) We conduct a temporal case study on historical newspapers from the Caribbean in the colonial period between 1770--1870. During this time, the region suffered both the consequences of European wars and political turmoil, as well as several uprisings of the local enslaved populations, which had a significant impact on the Caribbean social relationships and cultures \citep{migge:halshs-00674699}.
(3) To address the challenges of analysing historical documents, we probe the applied methods for their stability and ability to comprehend the noisy, archaic corpora.




We find that there is a trade-off between the stability of word embeddings and their compatibility with the historical dataset. Further, our temporal analysis connects changes in biased word associations to historical shifts taking place in the period. For instance, we couple the high association between \textit{Caribbean countries} and ``manual labour'' prevalent mostly in the earlier time periods to waves of white labour migrants coming to the Caribbean from 1750 onward.
Finally, we provide evidence supporting the intersectionality theory by observing conventional manifestations of gender bias solely for white people. While unsurprising, this finding necessitates intersectional bias analysis for historical documents.

\section{Related Work}

\paragraph{Intersectional Biases.}

Most prior work has analysed bias along one axis, e.g. race or gender, but not both simultaneously \citep{field-etal-2021-survey,stanczak-etal-2021-survey}. 
There, research on racial biases is generally centred around the gender majority group, such as Black men, while research on gender bias emphasises the experience of individuals who hold racial privilege, such as white women. Therefore, discrimination towards people with multiple minority identities, such as Black women, remains understudied. Addressing this, the intersectionality framework \citep{Crenshaw1989-CREDTI} investigates how different forms of inequality, e.g. gender and race, intersect with and reinforce each other. 
Drawing on this framework, \citet{tan-2019-assessing,may-etal-2019-measuring,lepori-2020-unequal,maronikolakis-etal-2022-analyzing,guo2021detecting} analyse the compounding effects of race and gender encoded in contextualised word representations and downstream tasks. Recently, \citet{lalor-etal-2022-benchmarking,jiang-fellbaum-2020-interdependencies} show the harmful implications of intersectionality effects in pre-trained language models.  
Less interest has been dedicated to unveiling intersectional biases prevalent in natural language, with a notable exception of \citet{kim2020intersectional} which provide evidence on intersectional bias in datasets of hate speech and abusive language on social media. As far as we know, this is the first paper on intersectional biases in historical documents.

\paragraph{Bias in Historical Documents.}

Historical corpora have been employed to study
societal phenomena such as language change \citep{kutuzov-etal-2018-diachronic,hamilton-etal-2016-diachronic} and societal biases. Gender bias has been analysed in biomedical research over a span of 60 years \citep{rios-etal-2020-quantifying}, in English-language books published between 1520 and 2008 \cite{hoyle-etal-2019-unsupervised}, and in Dutch newspapers from the second half of the 20th century \citep{wevers-2019-using}. 
\citet{sullam2022representation} investigate the evolution of the discourse on Jews in France during the 19th century. \citet{garg2018stereotypes} study the temporal change in stereotypes and attitudes  toward  women  and  ethnic  minorities  in  the  20th  and 21st  centuries in the US. However, they neglect the emergent intersectionality bias.  

When analysing the transformations of biases in historical texts, researchers rely on conventional tools developed for modern language. However, historical texts can be viewed as a separate domain due to their unique challenges of small and idiosyncratic corpora and noisy, archaic text \citep{piotrowski2012natural}.
Prior work has attempted to overcome the challenges such documents pose for modern tools, including recognition of spelling variations \citep{bollmann-2019-large} and misspelt words \citep{boros-etal-2020-alleviating}, and ensuring the stability of the applied methods
\citep{antoniak-mimno-2018-evaluating}. 

We study the dynamics of intersectional biases and their manifestations in language 
while addressing the challenges of historical data.

\section{Datasets}
\label{sec:data}

\begin{table}[t]
    \centering
    \fontsize{10}{10}\selectfont

    \begin{tabular}{lrr}
        \toprule
         Source & $\#$Files & $\#$Sentences \\ 
        \midrule 
        Caribbean Project & $7\,487$ & $5\,224\,591$  \\
        Danish Royal Library & $5\,661$ & $657\,618$  \\ \midrule 
        Total & $13\,148$ & $5\,882\,209$ \\
        \bottomrule
    \end{tabular}
    
    \caption{Statistics of the newspapers  dataset.}
    \label{tab:dataset_statistics}
\end{table}

\begin{table}[t]
    \centering
    \fontsize{10}{10}\selectfont

    \begin{tabular}{llrr}
        \toprule
        Period & Decade & $\#$Issues & Total  \\
        \midrule
        \multirow{4}{*}{}International & 1710--1770 & 15 & \multirow{4}{*}{$1\,886$} \\
        conflicts & 1770s &	747 & ~ \\ 
        and slave & 1780s &	283 & ~ \\ 
        rebellions & 1790s &	841 & ~ \\ 
        \midrule
        \multirow{3}{*}{}Revolutions & 1800s &	604  & \multirow{3}{*}{$3\,790$} \\ 
        and nation & 1810s &	$1\,347$ & ~ \\ 
        building & 1820s &	$1\,839$ & ~ \\ 
        \midrule
        \multirow{5}{*}{} & 1830s &	$1\,838$ &  \multirow{5}{*}{$7\,453$} \\  
        Abolishment & 1840s &	$1\,197$ & ~ \\ 
        of slavery & 1850s & $1\,111$ & ~ \\
        ~ & 1860s & $1\,521$ & ~ \\
        ~ & 1870s & $1\,786$ & ~ \\
        \bottomrule
    \end{tabular}
    
    \caption{Total number of articles in each period and decade.}
    \label{tab:datasets_periods}
\end{table}


         

Newspapers are considered 
an excellent source for the study of societal phenomena since they function as transceivers -- 
both producing and demonstrating public discourse \citep{wevers-2019-using}. As part of this study, we collect newspapers written in English from the  ``Caribbean Newspapers, 1718--1876'' database,\footnote{\url{https://www.readex.com/products/caribbean-newspapers-series-1-1718-1876-american-antiquarian-society}} the largest collection of Caribbean newspapers from the 18th--19th century available online. We extend this dataset with English-Danish newspapers published between 1770--1850 in the Danish colony of Santa Cruz (Saint Croix) downloaded from Danish Royal Library's website.\footnote{\url{https://www2.statsbiblioteket.dk/mediestream/}} See \Cref{tab:dataset_statistics} and \Cref{fig:caribbean_islands} (in \Cref{app:map}) for details.

As mentioned in \S\ref{sec:introduction}, the Caribbean islands experienced significant changes and turmoils during the 18th--19th century. 
Although chronologies can change from island to island, key moments in  Caribbean history can be divided into roughly four periods \citep{higman_2021,heuman2013caribbean}: 1) colonial trade and plantation system (1718 to 1750); 2) international conflicts and slave rebellions (1751 to 1790); 3) revolutions and nation building (1791 to 1825); 4) end of slavery and decline of European dominance (1826 to 1876). In our experimental setup, we conduct a temporal study on data split into these periods (see \Cref{tab:datasets_periods} 
for the number of articles in each period). As the resulting number of newspapers for the first period is very small ($<$ 10), we focus on the three latter periods. 


\begin{figure}[!t]
    \centering

        \includegraphics[width=\columnwidth, trim={0 1.2cm 6.5cm 0},clip]{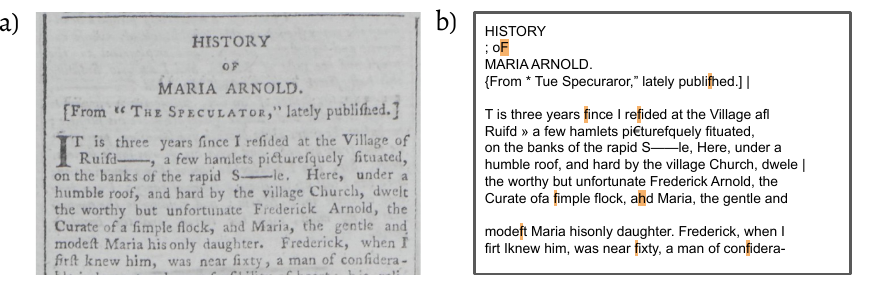}
         \caption{An example of a scanned newspaper (a) and the output of the OCR tool Tesseract (b). We fix simple OCR errors (highlighted) using a rule-based approach.}
         \label{fig:dataset_sample}
         
\end{figure}

\paragraph{Data Preprocessing.}

Starting with scans of entire newspaper issues (\Cref{fig:dataset_sample}.a), we first OCR them using the popular software Tesseract\footnote{\url{https://github.com/tesseract-ocr/tesseract}} with default parameters and settings. We then clean the dataset by applying the \texttt{DataMunging} package,\footnote{ \url{https://github.com/tedunderwood/DataMunging}} which uses a simple rule-based approach to fix basic OCR errors (e.g. long s' being OCRed as f', (\Cref{fig:dataset_sample}.b)). As some of the newspapers downloaded from the Danish royal library contain Danish text, we use \texttt{spaCy}\footnote{\url{https://spacy.io/}} to tokenise the OCRed newspapers into sentences and the python package \texttt{langdetect}\footnote{\url{https://github.com/Mimino666/langdetect}} to filter out non-English sentences.

\section{Bias and its Measures}
\label{sec:bias}
Biases can manifest themselves in natural language in many ways (see the surveys by \citet{stanczak-etal-2021-survey,field-etal-2021-survey,lalor-etal-2022-benchmarking}). In the following, we state the definition of bias we follow and describe the measures we use to quantify it.

\subsection{Definition}


Language is known to reflect common perceptions of the world \citep{hitti-etal-2019-proposed} and differences in its usage have been shown to reflect societal biases \citep{hoyle-etal-2019-unsupervised,marjanovic2022bias}. In this paper, we define bias in a text as the use of words or syntactic constructs that connote or imply an inclination or prejudice against a certain sensitive group, following the bias definition as in \citet{hitti-etal-2019-proposed}.
To quantify bias under this definition, we analyse 
word embeddings trained on our historical corpora.
These representations are assumed to carry lexical semantic meaning signals from the data and encode information about language usage in the proximity of entities. However, even words that are not used as direct descriptors of an entity influence its embedding, and thus its learnt meaning. 
Therefore, we further conduct an analysis focusing exclusively on words that describe identified entities.  



\subsection{Measures}
\label{sec:bias-measures}

\textbf{WEAT} The Word Embedding Association Test \citep{Caliskan_2017} is arguably the most popular benchmark to assess bias in word embeddings and has been adapted in numerous research \citep{may-etal-2019-measuring,rios-etal-2020-quantifying}.  
WEAT employs cosine similarity to measure the association between two sets of attribute words and two sets of target concepts. Here, the attribute words relate to a sensitive attribute (e.g. male and female), whereas the target concepts are composed of words in a category of a specific domain of bias (e.g. career- and family-related words). For instance, the WEAT statistic informs us whether the learned embeddings representing the concept of $family$ are more associated with females compared to males. 
According to \citet{Caliskan_2017}, the differential association between two 
sets of target concept embeddings, denoted $X$ and $Y$, with two sets of attribute embeddings, denoted as $A$ and $B$, can be calculated as: 

\begin{equation}
     s(X, Y, A, B) = \sum_{x \in X}\text{s}(x, A, B) - \sum_{y \in Y}\text{s}(y, A, B)
\nonumber
\end{equation}

\noindent where $s(w,A,B)$ measures the embedding association between one target word $w$ and each of the sensitive attributes:
\begin{equation}
    s(w, A, B) = \underset{a \in A}{\text{mean}}[\text{cos}(w,a)] - \underset{b \in B}{\text{mean}}[\text{cos}(w,b)]
\nonumber
\end{equation}

The resulting effect size is then a normalised measure of association:
\begin{equation}
    d = \frac{\underset{x \in X}{\text{mean}}[\text{s}(x,A,B)] - \underset{y \in Y}{\text{mean}}[\text{s}(y,A,B)]}{\underset{w \in X \cup Y}{\text{std}}[\text{s}(w, A, B)]}
\nonumber
\end{equation}

As a result, larger effect sizes imply a more biased word embedding. Furthermore, concept-related words should be equally associated with either sensitive attribute group assuming an unbiased word embedding. 






\noindent \textbf{PMI} We use point-wise mutual information (PMI; \citealt{church-hanks-1990-word}) as a measure of association between a descriptive word and a sensitive attribute (gender or race). In particular, PMI measures the difference between the probability of the co-occurrence of a word and an attribute, and their joint probability if they were independent as: 

\begin{equation}
    \text{PMI}(a,w)= \log \frac{p(a,w)}{p(a)p(w)}
\label{eq:pmi}
\end{equation}

A strong association with a specific gender or race leads to a high PMI. 
For example, a high value for $\text{PMI}(female, wife)$ is expected due to their co-occurrence probability being higher than the independent probabilities of $\mathit{female}$ and $\mathit{wife}$.
Accordingly, in an ideal unbiased world, words such as $\mathit{honourable}$ would have a PMI of approximately zero for all gender and racial identities.

\section{Experimental Setup}
\label{sec:experimental}
We perform two sets of experiments on our historical newspaper corpus. First, before we employ word embeddings to measure bias, we investigate the stability of the word embeddings trained on our dataset and evaluate their understanding of the noisy nature of the corpora. Second, we assess gender and racial biases using tools defined in \S\ref{sec:bias-measures}.

\subsection{Embedding Stability Evaluation}
\label{sec:exp-stability}

We use word embeddings as a tool to quantify historical trends and word associations in our data.   
However, prior work has called attention to the lack of stability of word embeddings trained on small and potentially idiosyncratic corpora \citep{antoniak-mimno-2018-evaluating,gonen-etal-2020-simple}.
We compare these different embeddings setups by testing them with regard to their stability and capturing meaning while controlling for the tokenisation algorithm, embedding size and the minimum number of occurrences.

We construct the word embeddings employing the continuous skip-gram negative sampling model from Word2vec \citep{mikolov2013word} using  \texttt{gensim}.\footnote{\url{https://radimrehurek.com/gensim/models/word2vec.html}} 
Following prior work \citep{antoniak-mimno-2018-evaluating,gonen-etal-2020-simple}, we test two common vector dimension sizes of 100 and 300, and two minimum numbers of occurrences of 20 and 100.
The rest of the hyperparameters are set to their default value.
We use two different methods for tokenising documents, the 
\texttt{spaCy} tokeniser and a subword-based tokeniser, Byte-Pair Encoding (BPE, \citet{gage1994bpe}). We train the BPE tokeniser on our dataset using the Hugging Face tokeniser implementation.\footnote{\url{https://huggingface.co/docs/tokenizers}} 

For each word in the vocabulary, we identify its 20 nearest neighbours and calculate the Jaccard similarity across five algorithm runs. Next, we test how well the word embeddings deal with the noisy nature of our documents. We create a list of 110 frequently misspelt words (See \Cref{app:amisspelt_Words}). We construct the list by first tokenising our dataset using \texttt{spaCy} and filtering out proper nouns and tokens that appear in the English dictionary. We then order the remaining tokens by frequency and manually scan the top $1\,000$ tokens for misspelt words. We calculate the percentage of words (averaged across 5 runs) for which the misspelt word is in immediate proximity to the correct word (top 5 nearest neighbours in terms of cosine similarity).

Based on the results of the stability and compatibility study, we select the most suitable model with which we conduct the following bias evaluation. 



\subsection{Bias Estimation}
\label{sec:exp-bias}

\subsubsection{WEAT Evaluation}
\label{sec:weat-evaluation}

As discussed in \S\ref{sec:bias-measures}, WEAT is used to evaluate how two attributes are associated with two target concepts in an embedding space, here of the model that was selected by the method described in \S\ref{sec:exp-stability}. 

In this work, we focus on the attribute pairs (\textit{female}, \textit{male})\footnote{As we deal with historical documents from the 18th--19th centuries, other genders are unlikely to be found in the data.} and (\textit{white}, \textit{non-white}). Usually, comparing the sensitive attributes (\textit{white}, \textit{non-white}) is done by collecting the embedding of popular white names and popular non-white names \cite{tan2019assessing}. However, this approach can introduce noise when applied to our dataset \citep{handler1996slave}. First, non-whites are less likely to be mentioned by name in historical newspapers compared to whites. Second, popular non-white names of the 18th and 19th centuries differ substantially from popular non-white names of modern times, and, to the best of our knowledge, there is no list of common historical non-white names.  For these reasons, instead of comparing the pair (\textit{white}, \textit{non-white}), we compare the pairs (\textit{African countries}, \textit{European countries}) and (\textit{Caribbean countries}, \textit{European countries}).

Following \citet{rios-etal-2020-quantifying}, we analyse the association of the above-mentioned attributes to the target concepts (\textit{career}, \textit{family}), (\textit{strong}, \textit{weak}), (\textit{intelligence}, \textit{appearance}), and (\textit{physical illness}, \textit{mental illness}). Following a consultation with a historian, we add further target concepts relevant to this period (\textit{manual labour}, \textit{non-manual labour}) and (\textit{crime}, \textit{lawfulness}). \Cref{tab:weat_keywords} (in \Cref{app:keyword_sets}) lists the target and attribute words we use for our analysis.

We also train a separate word embedding model on each of the dataset splits defined in \S\ref{sec:data} and run WEAT on the resulting three models. Comparing the obtained WEAT scores allows us to visualise temporal changes in the bias associated with the attributes and understand its dynamics.  

\subsubsection{PMI Evaluation}

Different from WEAT, calculating PMI requires first identifying entities in the OCRed historical newspapers and then classifying them into pre-defined attribute groups. The next step is collecting descriptors, i.e. words that are used to describe the entities. Finally, we use PMI to measure the association strength of the collected descriptors with each attribute group.

\paragraph{Entity Extraction.}
 We apply \texttt{F-coref} \cite{otmazgin-etal-2022-f}, a model for English coreference resolution that simultaneously performs entity extraction and coreference resolution on the extracted entities. The model's output is a set of entities, each represented as a list of all the references to that entity in the text. We filter out non-human entities by using \texttt{nltk}'s WordNet package,\footnote{\url{https://www.nltk.org/howto/wordnet.html}} retaining only entities for which the synset ``person.n1'' is a hypernym of one of their references.  

\label{sec:pmi-evaluation}

\begin{table*}[t]
    \centering
    \fontsize{10}{10}\selectfont

    \begin{tabular}{rrrrrr}
        \toprule
        $\#$Entities & $\#$Males & $\#$Females & $\#$Non-whites & $\#$Non-white males & $\#$Non-white females \\ 
        \midrule 
        $601\,468$  & $387\,292$ & $78\,821$ & $8\,525$ & $4\,543$ & $1\,548$\\
        \bottomrule
    \end{tabular}
    
    \caption{The entities in our \caribbeandataset  dataset. Notice that $\#$males and $\#$females do not sum to $\#$entities as some entities could not be classified. Similarly, $\#$non-white males and $\#$non-white females do not sum to $\#$non-whites.}
    \label{tab:entities_in_datasets}
\end{table*}

\paragraph{Entity Classification.} 
\label{sec:calssification}
We use a keyword-based approach \citep{lepori-2020-unequal} to classify the entities into groups corresponding to the gender and race axes and their intersection. Specifically, we classify each entity as being a member of \textit{male} vs \textit{female}, and \textit{white} vs \textit{non-white}. Additionally, entities are classified into intersectional groups (e.g. we classify an entity into the group \textit{non-white females} if it belongs to both \textit{female} and \textit{non-white}).    

Formally, we classify an entity $e$ with references $\{r^1_e, ..., r^m_e\}$ to attribute group $G$ with keyword-set $K_G=\{k_1,...,k_n\}$ if $\exists i$  such that $ r_e^i \in K_G$. See \Cref{app:keyword_sets} for listing the keyword sets of the different groups. In \Cref{tab:entities_in_datasets}, we present the number of entities classified into each group. We note here the unbalanced representation of the groups in the dataset. Further, it is important to state, that because it is highly unlikely that an entity in our dataset would be explicitly described as white, we classify an entity into the \textit{whites} group if it was not classified as \textit{non-white}. See \hyperref[sec:limitations]{Limitations} for a discussion of the limitations of using a keyword-based classification approach.

To evaluate our classification scheme, an author of this paper manually labelled a random sample of 56 entities. The keyword-based approach assigned the correct gender and race label for $\sim 80\%$ of the entities. See additional details in \Cref{tab:classification_acc} in \Cref{app:results}. From a preliminary inspection, it appears that many of the entities that were wrongly classified as \textit{female} were actually ships or other vessels (traditionally ``ship'' has been referred to using female gender). As \texttt{F-coref} was developed and trained using modern corpora, we evaluate its accuracy on the same set of 56 entities. Two authors of this paper validated its performance on the historical data to be satisfactory, with especially impressive results on shorter texts with fewer amount of OCR errors.

\paragraph{Descriptors Collection.}
Finally, we use \texttt{spaCy} to collect descriptors for each classified entity. Here, we define the descriptors as the lemmatised form of tokens that share a dependency arc labelled ``amod'' (i.e. adjectives that describe the tokens) to one of the entity's references. Every target group $G_j$ is then assigned with descriptors list $D_j = [d_1, ..., d_{k}]$. 


To calculate PMI according to \Cref{eq:pmi}, we estimate the joint distribution of a target group and a descriptor using a simple plug-in estimator:
\begin{align}
    \widehat{p}(G_j, d_i) \propto \mathrm{count}(G_j, d_i)
\end{align}

\noindent Now, we can assign every word $d_i$ two continuous values representing its bias in the gender and race dimensions by calculating $\text{PMI}(\textit{female},d_i) - \text{PMI}(\textit{males},d_i)$ and $\text{PMI}(\textit{non-white},d_i) - \text{PMI}(\textit{white},d_i)$. These two continuous values can be seen as $d_i$'s coordinates on the intersectional gender/race plane.

\begin{table}[ht]
\centering
\fontsize{10}{10}\selectfont
 \begin{tabular}{lrr|rrr}
    \rotatebox{90}{Tokenisation} & \rotatebox{90}{\parbox{2cm}{Embedding \\ Size}} & \rotatebox{90}{Min Freq} & \rotatebox{90}{\parbox{2cm}{Mean JS \\ Top 20}} & \rotatebox{90}{\parbox{2cm}{Correct Word \\ in Top 5 \\ (all words)}}  & \rotatebox{90}{\parbox{2cm}{\% Misspelling \\ in vocabulary}} \\ \midrule
    BPE & 100 & 20 & \textbf{0.66} & 37.04 & 94.44 \\ 
     & 100 & 100 & \textbf{0.66} & 37.04 & 94.44\\
     & 300 & 20 & 0.63 & 40.74 & 94.44\\ 
     & 300 & 100 & 0.64 & 39.81 & 94.44\\ \midrule
    SpaCy & 100 & 20 & 0.59 & \textbf{63.89}& 74.07\\
    & 100 & 100 & 0.65 & 48.15& 56.48\\
     & 300 & 20 & 0.55 & \textbf{63.89}&74.07\\
     & 300 & 100 & 0.61 & 50.00& 56.48\\
    \bottomrule %

 \end{tabular}
 \caption{Results of the stability analysis of different word embedding methods (measured with Jaccard similarity) and their compatibility with the historical corpora (ability to recognise misspelt words).}
 \label{tab:embedding_methods}
\end{table}

\begin{figure*}[ht]
    \centering

        \includegraphics[width=\textwidth, trim={0cm 8.6cm 3.2cm 0cm},clip]{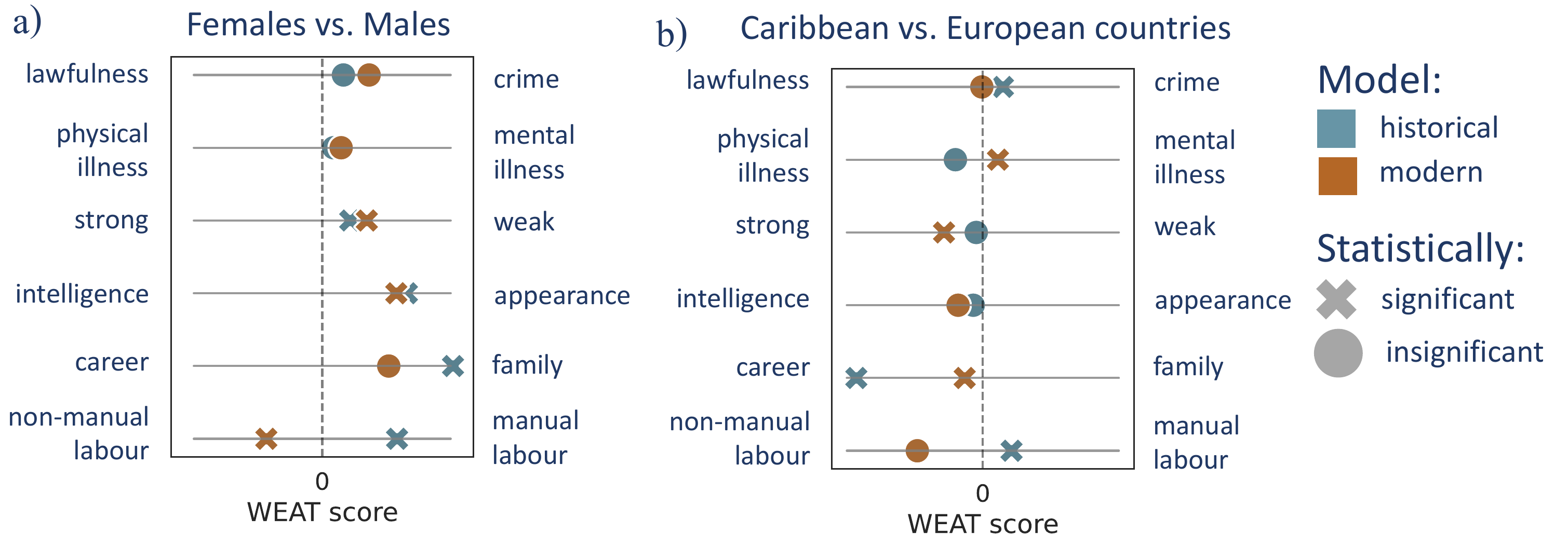}
         \caption{a) WEAT results of \textit{females} vs \textit{males}. The location of a marker measures the association strength of \textit{females} with the concept (compared to \textit{males}). For example, according to the modern model, \textit{females} are associated with ``weak'' and \textit{non-manual labour} while \textit{males} are associated with ``strong'' and \textit{manual labour}. b) WEAT results of \textit{Caribbean countries} vs \textit{European countries}. The location of a marker measures the association strength of \textit{Caribbean countries} with the concept (compared to \textit{European countries}).}
         \label{fig:weat_all}
         
\end{figure*}

\begin{figure*}[ht]
    \centering

        \includegraphics[width=\textwidth, trim={0cm 0cm 0.0cm 0.0cm},clip]{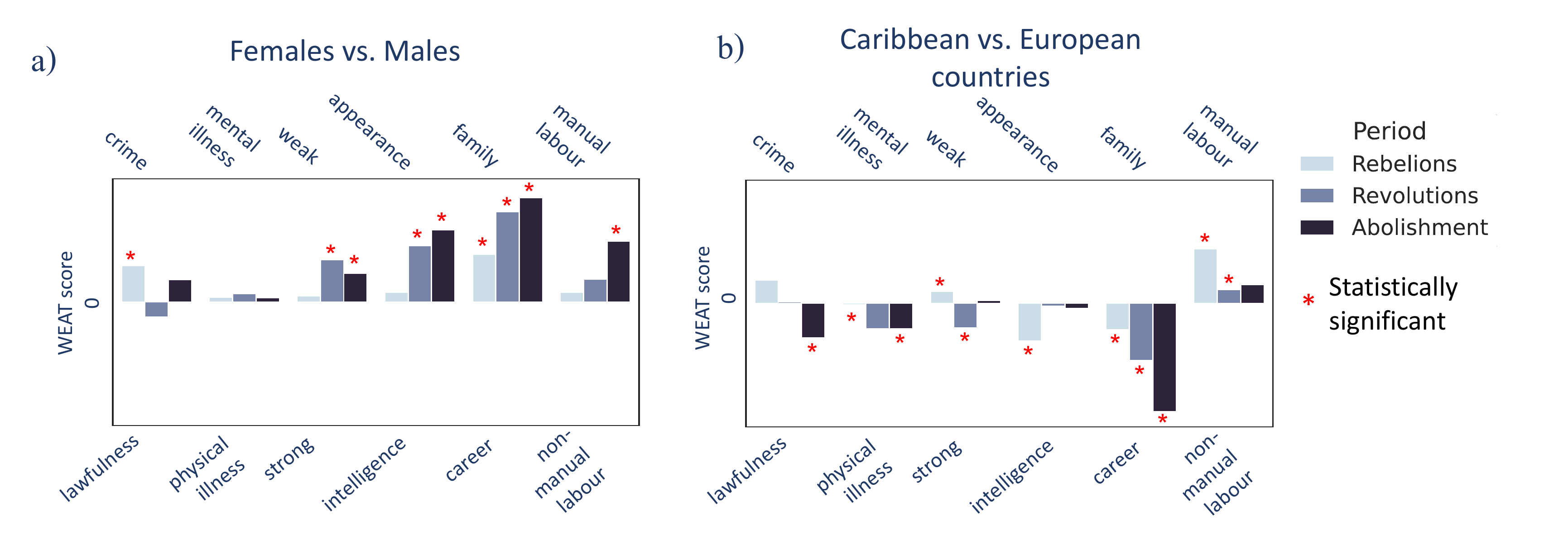}
         \caption{Temporal WEAT analysis conducted for the periods 1751--1790 (rebellions), 1791--1825 (revolutions) and 1826--1876 (abolishment). Similar to \Cref{fig:weat_all}, the height of each bar represents how strong the association of the attribute is with each concept.}
         \label{fig:weat_temporal}
         
\end{figure*}

\subsubsection{Lexicon Evaluation}

Another popular approach for quantifying different aspects of bias is the application of specialised lexica \citep{stanczak-etal-2021-survey}.
These lexica assign words a continuous value that represents how well the word aligns with a specific dimension of bias. We use NRC-VAD lexicon \citep{mohammad-2018-obtaining} to compare word usage associated with the sensitive attributes \textit{race} and \textit{gender} in three dimensions: \textit{dominance} (strength/weakness), \textit{valence} (goodness/badness), and \textit{arousal} (activeness/passiveness of an identity). Specifically, given a bias dimension $\mathcal{B}$ with lexicon ${L_\mathcal{B}} = \{(w_1, a_1), ..., (w_n, a_n)\}$, where $(w_i, a_i)$ are word-value pairs, we calculate the association of $\mathcal{B}$ with a sensitive attribute $G_j$ using:

\begin{equation}
    \begin{split}
        A(\mathcal{B}, G_j)  = \frac{\sum_i^n a_i \cdot \text{count}(w_i, D_j)}{\sum_i^n \text{count}(w_i, D_j)} 
    \end{split}
\end{equation}

\noindent where $\text{count}(w_i, D_j)$ is the number of times the word $w_i$ appears in the descriptors list $D_j$.

%




        
    

\section{Results}
\label{sec:results}

First, we investigate 
which training strategies of word embeddings optimise their stability and compatibility on historical corpora (\S\ref{sec:result-stability}). Next, we analyse 
how bias is manifested along the gender and racial axes and whether there are any
noticeable differences in bias across different periods of the Caribbean history (\S\ref{sec:result-bias}). 

\subsection{Embedding Stability Evaluation}
\label{sec:result-stability}

In \Cref{tab:embedding_methods}, we present the results of the study on the influence of training strategies of word embeddings. 
We find that there is a trade-off between the stability of word embeddings and their compatibility with the dataset. 
While BPE achieves a higher Jaccard similarity across the top 20 nearest neighbours for each word across all runs, it loses the meaning of misspelt words. Interestingly, this phenomenon arises, despite the misspelt words occurring frequently enough to be included in the BPE model's vocabulary. 

For the remainder of the experiments, we aim to select a model which effectively manages this trade-off achieving both high stability and captures meaning despite the noisy nature of the underlying data. 
Thus, we opt to use a \texttt{spaCy}-based embedding with a minimum number of occurrences of 20 and an embedding size of 100 which achieves competitive results in both of these aspects. 
Finally, we note that our results remain stable across different algorithm runs and do not suffer from substantial variations which corroborates the reliability of the findings we make henceforth.


         


         

\begin{figure}[t]
    \centering

        \includegraphics[width=\columnwidth, trim={0cm 0cm 0cm 0cm},clip]{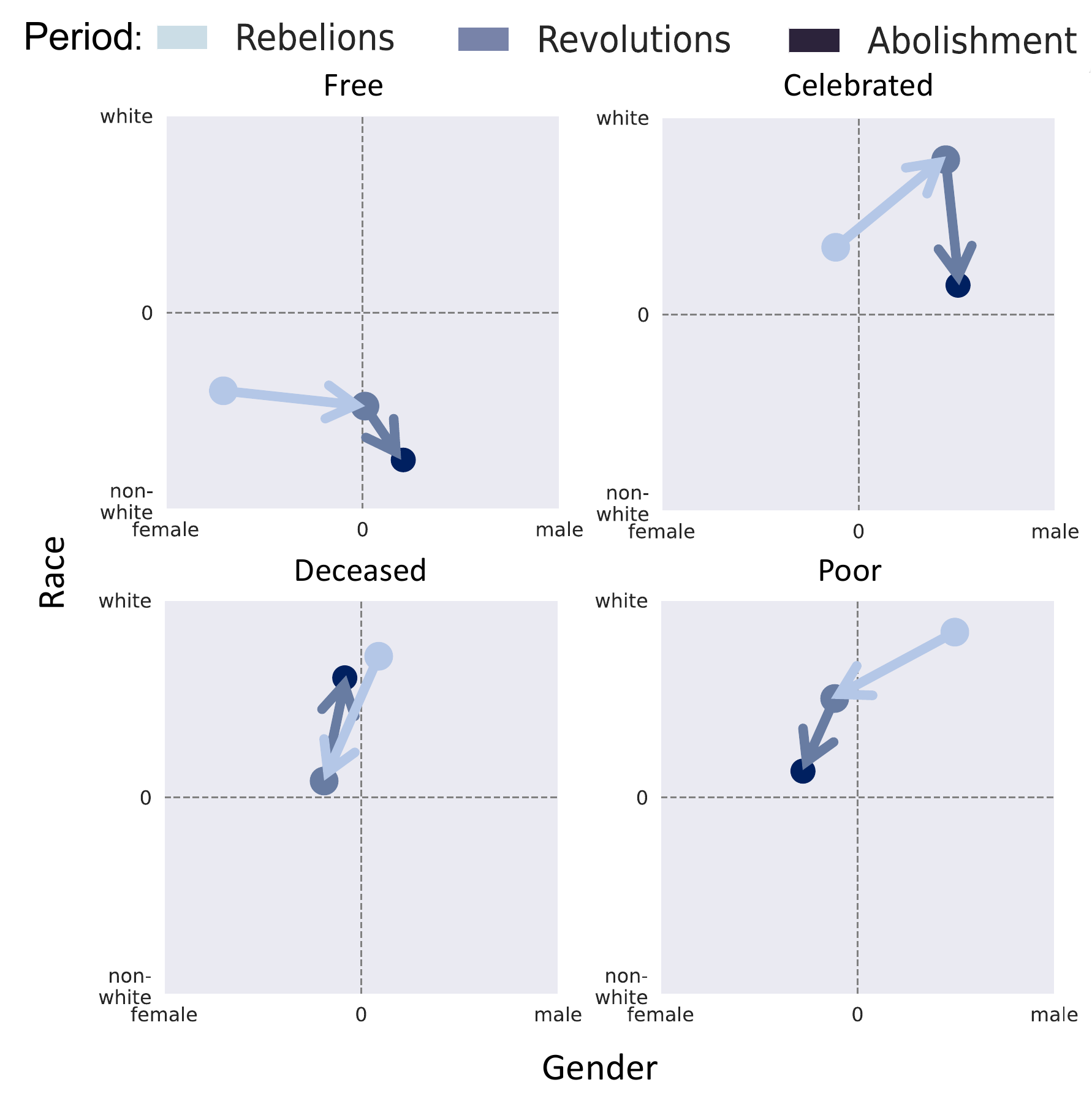}
         \caption{Intersectional PMI analysis of ``free'', ``celebrated'', ``deceased'' and ``poor'' across the periods.}
         \label{fig:temporal_pmi}
         
\end{figure}

\subsection{Bias Estimation}
\label{sec:result-bias}

\subsubsection{WEAT Analysis}
\label{sec:weat_Analysis}

\Cref{fig:weat_all} displays the results of performing a WEAT analysis for measuring the association of the six targets described in \S\ref{sec:exp-bias} with the attributes (\textit{females}, \textit{males}) and (\textit{Caribbean countries}, \textit{European countries}), respectively.\footnote{See \Cref{fig:weat_3} in \Cref{app:results} for analysis of the attributes (\textit{African countries}, \textit{European countries}).} We calculate the WEAT score using the embedding model from \S\ref{sec:result-stability} and compare it with an embedding model trained on modern news corpora (\texttt{word2vec-google-news-300}, \citet{mikolov2013efficient}). We notice interesting differences between the historical and modern embeddings. For example, while in our dataset \textit{females} are associated with the target concept of \textit{manual labour}, this notion is more aligned with \textit{males} in the modern corpora. A likely cause is that during this period, womens' intellectual and administrative work was not commonly recognised \cite{wayne2020women}.
It is also interesting to note that the attribute \textit{Caribbean countries} has a much stronger association in the historical embedding with the target \textit{career} (as opposed to \textit{family}) compared to the modern embeddings. A possible explanation is that Caribbean newspapers referred to locals by profession or similar titles, while Europeans were referred to as relatives of the Caribbean population.

In \Cref{fig:weat_temporal} and \Cref{fig:weat_temp_africa} (in \Cref{app:results}), we present a dynamic WEAT analysis that unveils trends on a temporal axis. In particular, we see an increase in the magnitude of association between the target of \textit{family} vs \textit{career} and the attributes (\textit{females}, \textit{males}) and (\textit{Caribbean countries}, \textit{European countries}) over time. It is especially interesting to compare \Cref{fig:weat_all} with \Cref{fig:weat_temporal}. One intriguing result is that the high association between \textit{Caribbean countries} and \textit{manual labour} can be attributed to the earlier periods.  This finding is potentially related to several historical shifts taking place in the period. For instance, while in the earlier years, it was normal for plantation owners to be absentees and continue to live in Europe, from 1750 onward, waves of white migrants with varied professional backgrounds
came to the Caribbean.


\begin{figure}[!t]
    \centering
        \includegraphics[width=\columnwidth, trim={0.25cm 0 0 0},clip]{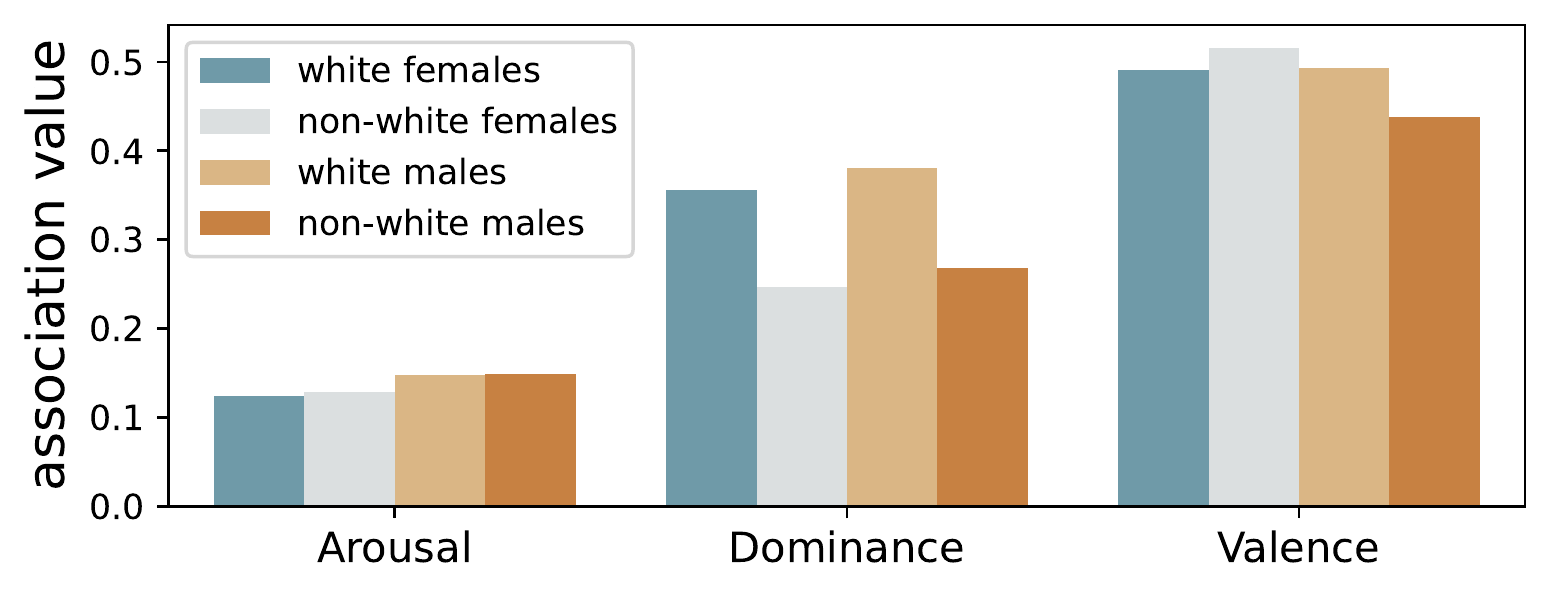}
         \caption{Association of attributes with the lexicon of dominance, valence, and arousal.}
         \label{fig:lexicon_results}
\end{figure}

\begin{figure*}[t]
    \centering

        \includegraphics[width=\textwidth, trim={0.25cm 5cm 0.2cm 4cm},clip]{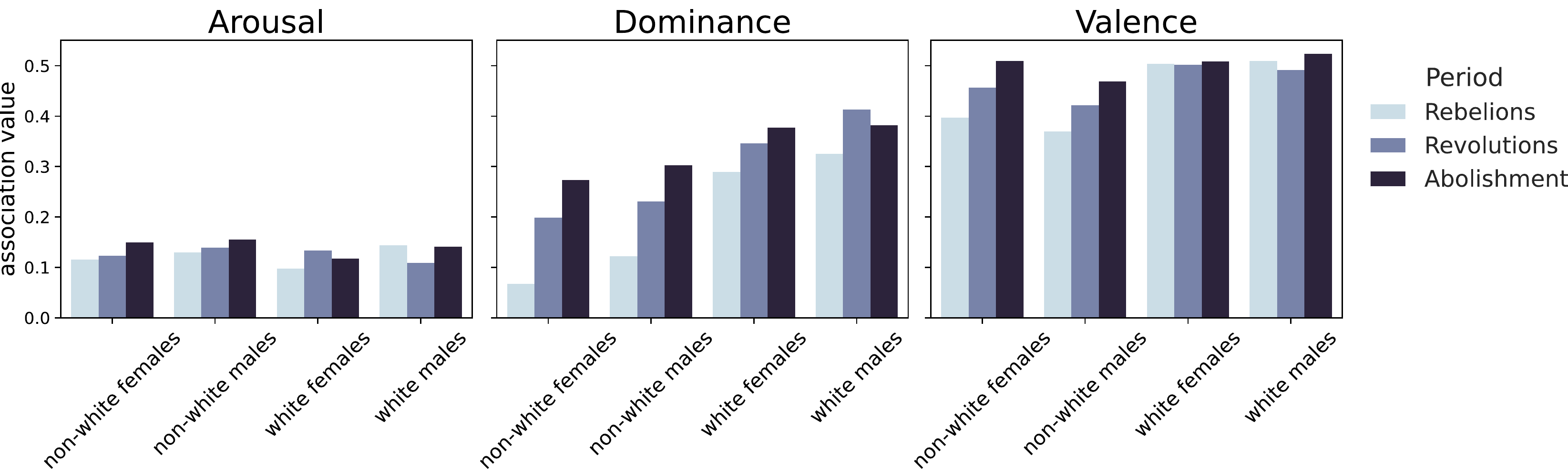}
         \caption{Association of attributes with the lexicon of dominance, valence, and value done on the periods 1751--1790 (rebellions), 1791--1825 (revolutions) and 1826--1876 (abolishment).}
         \label{fig:lexicon_periods}
         
\end{figure*}

\subsubsection{PMI Analysis}

We report the results of the intersectional PMI analysis in \Cref{fig:general_bias_descriptors}. As can be seen, an intersectional analysis can shed a unique light on the biased nature of some words in a way that single-dimensional analysis cannot. \textit{White males} are ``brave'' and ``ingenious'', and \textit{non-white males} are described as ``active'' and ``tall''. Interestingly, while words such as ``pretty'' and ``beautiful'' (and peculiarly, ``murdered'') are biased towards \textit{white} as opposed to \textit{non-white females}, the word ``lovely'' is not, whereas ``elderly'' is strongly aligned with \textit{non-white females}. Another intriguing dichotomy is the word pair ``sick'' and ``blind'' which are both independent along the gender axis but manifest a polar racial bias. In \Cref{tab:interesting_examples} in \Cref{app:results}, we list some examples from our dataset featuring those words.     

Similarly to \S\ref{sec:weat_Analysis}, we perform a temporal PMI analysis by comparing results obtained from separately analysing the three dataset splits. In \Cref{fig:temporal_pmi}, we follow the trajectory over time of the biased words ``free'', ``celebrated'', ``deceased'' and ``poor''. Each word displays different temporal dynamics. For example, while the word ``free'' moved towards the \textit{male} attribute, ``poor'' transitioned to become more associated with the attributes \textit{female} and \textit{non-white} over time (potentially due to its meaning change from an association with poverty to a pity).

These results provide evidence for the claims of the intersectionality theory. We observe conventional manifestations of gender bias, i.e. ``beautiful'' and ``pretty'' for \textit{white females}, and ``ingenious'' and ``brave'' for \textit{white males}. While unsurprising due to the societal status of non-white people in that period, this finding necessitates intersectional bias analysis for historical documents in particular.

\subsubsection{Lexicon Evaluation}

Finally, we report the lexicon-based evaluation results in \Cref{fig:lexicon_results} and \Cref{fig:lexicon_periods}. Unsurprisingly, we observe lower dominance levels for the \textit{non-white} and \textit{female} attributes compared to \textit{white} and \textit{male}, a finding previously uncovered in modern texts \citep{field-tsvetkov-2019-entity,rabinovich-etal-2020-pick}. While \Cref{fig:lexicon_periods} indicates that the level of dominance associated with these attributes raised over time, a noticeable disparity to white males remains. Perhaps more surprising is the valence dimension. We see the highest and lowest levels of associations with the intersectional attributes \textit{non-white female} and \textit{non-white male}, respectively. We hypothesise that this connects to the nature of advertisements for lending the services of or selling non-white women where being agreeable is a valuable asset.

\section{Conclusions}
\label{sec:conclusion}
In this paper, we examine biases present in historical newspapers published in the Caribbean during the colonial era by conducting a temporal analysis of biases along the axes of gender, race, and their intersection. We evaluate the effectiveness of different embedding strategies and find a trade-off between the stability and compatibility of word representations on historical data. 
We link changes in biased word usage to historical shifts, coupling the development of the association between \textit{manual labour} and \textit{Caribbean countries} to waves of white labour migrants coming to the Caribbean from 1750 onward. Finally, we provide evidence to corroborate the intersectionality theory by observing conventional manifestations of gender bias solely for white people.  


\section*{Limitations}
\label{sec:limitations}

We see several limitations regarding our work. First, we focus on documents in the English language only, neglecting many Caribbean newspapers and islands with other official languages. While some of our methods can be easily extended to non-English material (e.g. WEAT analysis), methods that rely on the pre-trained English model \texttt{F-coref} (i.e. PMI, lexicon-based analysis) can not. 

On the same note, \texttt{F-coref} and \texttt{spaCy} were developed and trained using modern corpora, and their capabilities when applied to the noisy historical newspapers dataset, are noticeably lower compared to modern texts. Contributing to this issue is the unique, sometimes archaic language in which the newspapers were written. While we validate \texttt{F-coref} performance on a random sample (\S\ref{sec:exp-bias}), this is a significant limitation of our work. Similarly, increased attention is required to adapt the keyword sets used by our methods to historical settings.

Moreover, our historical newspaper dataset is inherently imbalanced and skewed. As can be seen in \Cref{tab:datasets_periods} and \Cref{fig:caribbean_islands}, there is an over-representation of a handful of specific islands and time periods. While it is likely that in different regions and periods, less source material survived to modern times, part of the imbalance (e.g. the prevalence of the US Virgin Islands) can also be attributed to current research funding and policies.\footnote{The Danish government has recently funded a campaign for the digitisation of historical newspapers published in the Danish colonies; \url{https://stcroixsource.com/2017/03/01/}.} Compounding this further, minority groups are traditionally under-represented in news sources. This introduces noise and imbalance into our results, which rely on a large amount of textual material referring to each attribute on the gender/race plane that we analyse.

Relating to that, our keyword-based method of classifying entities into groups corresponding to the gender and race axes is limited. While we devise a specialised keyword set targeting the attributes \textit{female}, \textit{male} and \textit{non-white}, we classify an entity into the \textit{white} group if it was not classified as \textit{non-white}. This discrepancy is likely to introduce noise into our evaluation, as can also be observed in \Cref{tab:classification_acc}. This tendency may be intensified by the NLP systems that we use, as many tend to perform worse on gender- and race-minority groups \citep{field-etal-2021-survey}.

Finally, in this work, we explore intersectional bias only along the race and gender axes. Thus, we neglect the effects of other confounding factors (e.g. societal position, occupation) that affect asymmetries in language.

\section*{Ethical Considerations}

Studying historical texts from the era of colonisation and slavery poses ethical issues to historians and computer scientists alike since vulnerable groups still suffer the consequences of this history in the present. Indeed, racist and sexist language is not only a historical artefact of bygone days but has a real impact on people's lives \citep{alim_oxford_2020}.

We note that the newspapers we consider for this analysis were written foremost by the European oppressors. Moreover, only a limited number of affluent people (white males) could afford to place advertisements in those newspapers (which constitute a large portion of the raw material). This skews our study toward language used by privileged individuals and their perceptions.     

This work aims to investigate racial and gender biases, as well as their intersection. Both race and gender are considered social constructs and can encompass a range of perspectives, including one's reflected, observed, or self-perceived identity. In this paper, we classify entities as observed by the author of an article and infer their gender and race based on the pronouns and descriptors used in relation to this entity. We follow this approach in an absence of explicit demographic information. However, we warn that this method poses a risk of misclassification. Although the people referred to in the newspapers are no longer among the living, we should be considerate when conducting studies addressing vulnerable groups.  

Finally, we use the mutually exclusive \textit{white} and \textit{non-white} race categories as well as \textit{male} and \textit{female} gender categories. We acknowledge that these groupings do not fully capture the nuanced nature of bias. This decision was made due to limited data discussing minorities in our corpus.
While gender identities beyond the binary are unlikely to be found in the historical newspapers from the 18th-19th century, future work will aim to explore a wider range of racial identities.




\bibliography{anthology,custom}
\bibliographystyle{acl_natbib}

\clearpage
\appendix
\label{sec:appendix}

\section{Additional Material}
\label{app:additional_material}

\subsection{Dataset Statistics}
\label{app:map}

\begin{figure*}[t]
    \centering

        \includegraphics[width=\textwidth, trim={0 0 0 0},clip]{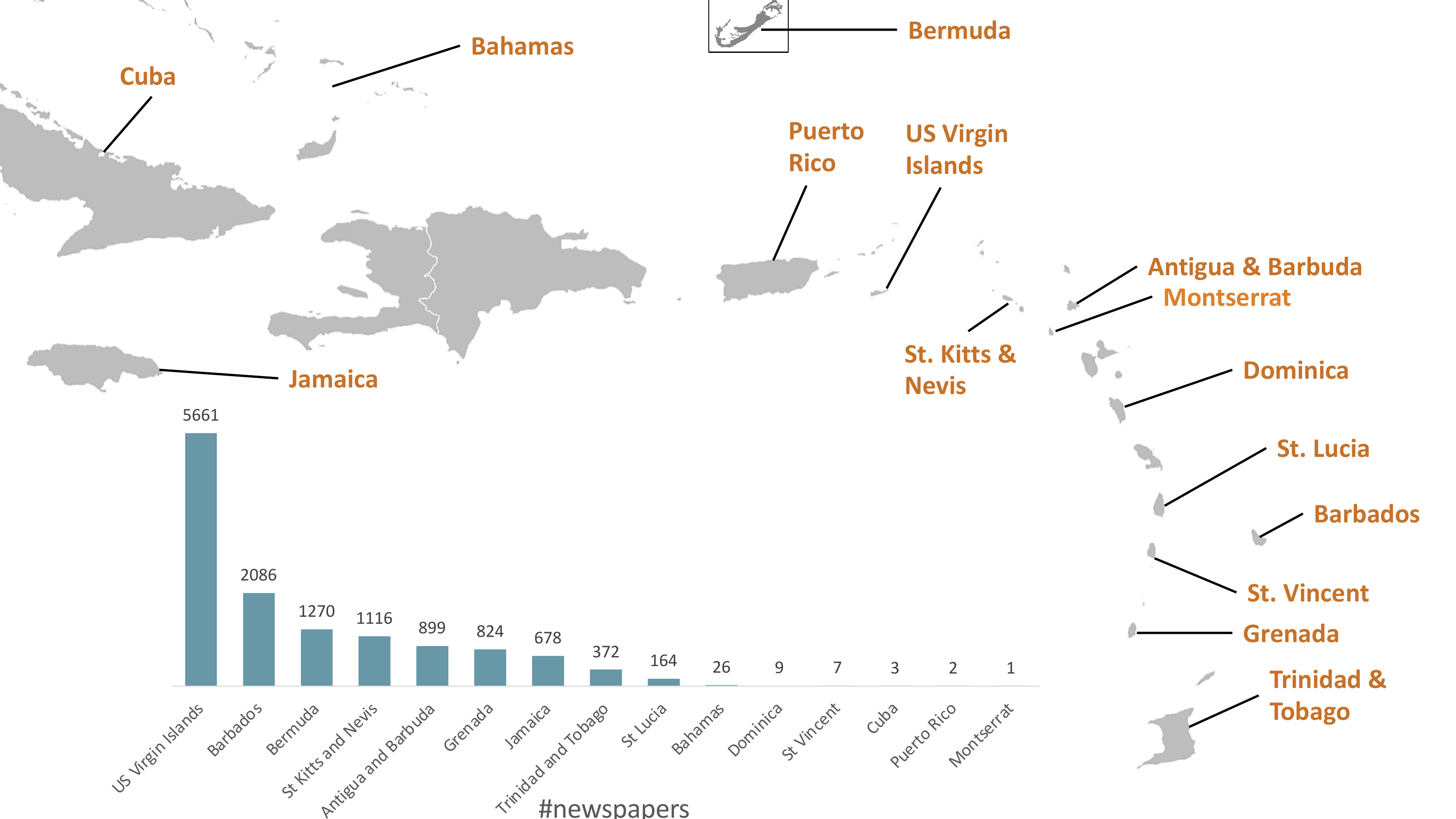}
         \caption{The geographical distribution of the curated Caribbean newspapers dataset.}
         \label{fig:caribbean_islands}
         
\end{figure*}

In \Cref{fig:caribbean_islands}, we present the geographical distribution of the newspapers in the curated dataset. 

\subsection{Misspelt Words}
\label{app:amisspelt_Words}

Here we list 110 frequently misspelt words and their correct spelling, which was used for the embedding evaluation described in \Cref{sec:exp-stability}.

hon’ble - honorable, honble - honorable, majetty - majesty, mujesty - majesty, mojesty - majesty, houfe - house, calied - called, upen - upon, cailed - called, reeeived - received, betore - before, kaow - know, reecived - received, bope - hope, fonnd - found, dificult - difficult, qnite - quite, convineed - convinced, satistied - satisfied, intinate - intimate, demandcd - demanded, snecessful - successful, abie - able, impossibie - impossible, althouch - although, foreed - forced, giad - glad, preper - proper, understocd - understood, fuund - found, almest - almost, nore - more, atter - after, oceupied - occupied, understuod - understood, satis'y - satisfy, impofible - impossible, impoilible - impossible, inseusible - insensible, accessary - accesory, contident - confident, koown - known, receiv - receive, calied - calles, appellunt - appellant, Eniperor - emperor, auxious - anxious, ofien - often, lawiul - lawful, posstble - possible, Svanish - Spanish, fuffictent - sufficient, furcher - further, yery - very, uader - under, ayreeable - agreeable, ylad - glad, egreed - agreed, unabie - unable, giyen - given, uecessary - necessary, alrendy - already, entitied - entitled, cffered - offered, pesitive - positive, creater - creator, prefound - profound, examived - examined, successiul - successful, pablic - public, propor - proper, cousiderable - considerable, lcvely - lovely, fold - sold, seeond - second, huuse - house, excellen - excellent, auetion - auction, Engiand - England, peopie - people, goveroment - government, yeurs - years, exceliency - excellency, generel - general, foliowing - following, goneral - general, preperty - property, wondertul - wonderful, o’ciock - o’clock, exeellency - excellency, tollowing - following, Eugland - England, gentieman - gentleman, colontal - colonial, gevernment - government, excelleney - excellency, goverament - government, Lendon - London, Bermupa - Bermuda, goverument - government, himeelf - himself, entlemen - gentlemen, sublcriber - subscriber, majeliy - majesty, Weduesday - Wednesday, o’cleck - o’clock, o’cluck - o’clock, colonics - colonies, sngar - sugar.

\subsection{Keyword Sets}
\label{app:keyword_sets}

\Cref{tab:classification_keywords} and \Cref{tab:weat_keywords} describe the various keyword sets that we used for entity classification (Section \ref{sec:calssification}) and for performing the WEAT tests (Section \ref{sec:weat-evaluation}. 

\begin{table*}[t]
\centering
\fontsize{10}{10}\selectfont
 \begin{tabular}{p{3cm}p{10cm}}
    \toprule
    Subgroup & Wordlist \\ \midrule
    Males & husband, suitor, brother, boyhood, beau, salesman, daddy, man, spokesman, chairman, lad, mister, men, sperm, dad, gelding, gentleman, boy, sir, horsemen, paternity, statesman, prince, sons, countryman, pa, suitors, stallion, fella, monks, fiance, chap, uncles, godfather, bulls, males, grandfather, penis, lions, nephew, monk, countrymen, grandsons, beards, schoolboy, councilmen, dads, fellow, colts, mr, king, father, fraternal,baritone, gentlemen, fathers, husbands, guy, semen, brotherhood, nephews, lion, lads, grandson, widower, bachelor, kings, male, son, brothers, uncle, brethren, boys, councilman, czar, beard, bull, salesmen, fraternity, dude, colt, john, he, himself, his \\ \midrule
    
    Females & sisters, queen, ladies, princess, witch, mother, nun, aunt, princes, housewife, women, convent, gals, witches, stepmother, wife, granddaughter, mis, widows, nieces, studs, niece, actresses, wives, sister, dowry, hens, daughters, womb, monastery, ms, misses, mama, mrs, fillies, woman, aunts, girl, actress, wench, brides, grandmother, stud, lady, female, maid, gal, queens, hostess, daughter, grandmothers, girls, heiress, moms, maids, mistress, mothers, mom, mare, filly, maternal, bride, widow, goddess, diva, maiden, hen, housewives, heroine, nuns, females', she, herself, hers, her \\ \midrule

    Non-whites & negro, negros, creole, indian, negroes, colored, mulatto, mulattos, negresse, mundingo, brown, browns, african, congo, black, blacks, dark, creoles \\ \midrule
    Whites & (any entity that was not classified as Non-white) \\
    \bottomrule %

 \end{tabular}
 \caption{Keywords used for classification entities into subgroups.}
 \label{tab:classification_keywords}
\end{table*}

\begin{table*}[t]
\centering
\fontsize{10}{10}\selectfont
 \begin{tabular}{p{2cm}p{14cm}}
    \toprule
    \textbf{Attribute} &\textbf{ Wordlist} \\ \midrule
    Males & husband, man, mister, gentleman, boy, sir, prince, countryman, fiance, godfather, grandfather, nephew, fellow, mr, king, father, guy, grandson, widower, bachelor, male, son, brother, uncle, brethren \\ \midrule
    
    Females & sister, queen, lady, witch, mother, aunt, princes, housewife, stepmother, wife, granddaughter, mis, niece, ms, misses, mrs, woman, girl, wench, bride, grandmother, female, maid, daughter, mistress, bride, widow, maiden \\ \midrule

    European countries & ireland, georgia, france, monaco, poland, cyprus, greece, hungary, norway, portugal, belgium, luxembourg, finland, albania, germany, netherlands, montenegro, scotland, spain, europe, russia, vatican, switzerland, lithuania, bulgaria, wales, ukraine, romania, denmark, england, italy, bosnia, turkey, malta, iceland, austria, croatia, sweden, macedonia \\ \midrule

    African countries & liberia, mozambique, gambia, ghana, morocco, chad, senegal, togo, algeria, egypt, benin, ethiopia, niger, madagascar, guinea, mauritius, africa, mali, congo, angola \\ \midrule

    Caribbean countries & barbuda, bahamas, jamaica, dominica, haiti, antigua, grenada, caribbean, barbados,  cuba, trinidad, dominican, nevis, kitts, lucia, croix, tobago, grenadines, puerto, rico \\
    \midrule \midrule
    \textbf{Target} & \textbf{Wordlist} \\ \midrule

    Appearance & apt, discerning, judicious, imaginative, inquiring, intelligent, inquisitive, wise, shrewd, logical, astute, intuitive, precocious, analytical, smart, ingenious, reflective, inventive, venerable, genius, brilliant, clever, thoughtful \\ \midrule

    Intelligence & bald, strong, muscular, thin, voluptuous, blushing, athletic, gorgeous, handsome, homely, feeble, fashionable, attractive, weak, plump, ugly, slim, stout, pretty, fat, sensual, beautiful, healthy, alluring, slender \\ \midrule

    Weak & failure, loser, weak, timid, withdraw, follow, fragile, afraid, weakness, shy, lose, surrender, vulnerable, yield  \\ \midrule

    Strong & strong, potent, succeed, loud, assert, leader, winner, dominant, command, confident, power, triumph, shout, bold  \\ \midrule

    Family & loved, sisters, mother, reunited, estranged, aunt, relatives, grandchildren, godmother, kin, grandsons, sons, son, parents, stepmother, childless, paramour, nieces, children, niece, father, twins, sister, fiance, daughters, youngest, uncle, uncles, aunts, eldest, cousins, grandmother, children, loving, daughter, paternal, girls, nephews, friends, mothers, grandfather, cousin, maternal, married, nephew, wedding, grandson \\ \midrule

    Career & branch, managers, usurping, subsidiary, engineering, performs, fiscal, personnel, duties, offices, clerical, engineer, executive, functions, revenues, entity, competitive, competitor, employing, chairman, director, commissions, audit, promotion, professional, assistant, company, auditors, oversight, departments, comptroller, president, manager, operations, marketing, directors, shareholder, engineers,  corporate, salaries, internal, management, salaried, corporation, revenue, salary, usurpation, managing, delegated, operating  \\ \midrule

    Manual labour & sailor, bricklayer, server, butcher, gardener, cook, repairer, maid, guard, farmer, fisher, carpenter, paver, cleaner, cabinetmaker, barber, breeder, washer, miner, builder, baker, fisherman, plumber, labourer, servant \\ \midrule

    Non-manual labour & teacher, judge, manager, lawyer, director, mathematician, physician, medic, designer, bookkeeper, nurse, librarian, doctor, educator, auditor, clerk, midwife, translator, inspector, surgeon \\ \midrule

    Mental illness & sleep, pica, disorders, nightmare, personality, histrionic, stress, dependence, anxiety, terror, emotional, delusion, depression, panic, abuse, disorder, mania, hysteria \\ \midrule 

    Physical illness & scurvy, sciatica, asthma, gangrene, gerd, cowpox, lice, rickets, malaria, epilepsy, sars, diphtheria, smallpox, bronchitis, thrush, leprosy, typhus, sids, watkins, measles, jaundice, shingles, cholera, boil, pneumonia, mumps, rheumatism, rabies, abscess, warts, plague, dysentery, syphilis, cancer, influenza, ulcers, tetanus \\ \midrule

    Crime & arrested, unreliable, detained, arrest, detain, murder, murdered, criminal, criminally, thug, theft, thief, mugger, mugging, suspicious, executed, illegal, unjust, jailed, jail, prison, arson, arsonist, kidnap, kidnapped, assaulted, assault, released, custody, police, sheriff, bailed, bail \\ \midrule 

    lawfulness & loyal, charming, friendly, respectful, dutiful, grateful, amiable, honourable, honourably, good, faithfully, faithful, pleasant, praised, just, dignified, approving, approve, compliment, generous, faithful, intelligent, appreciative, delighted, appreciate \\
    \bottomrule %

 \end{tabular}
 \caption{Keywords used for performing WEAT evaluation.}
 \label{tab:weat_keywords}
\end{table*}

\section{Supplementary Results}
\label{app:results}

\begin{figure}[ht]
    \centering

        \includegraphics[width=\columnwidth, trim={0cm 12cm 5cm 0cm},clip]{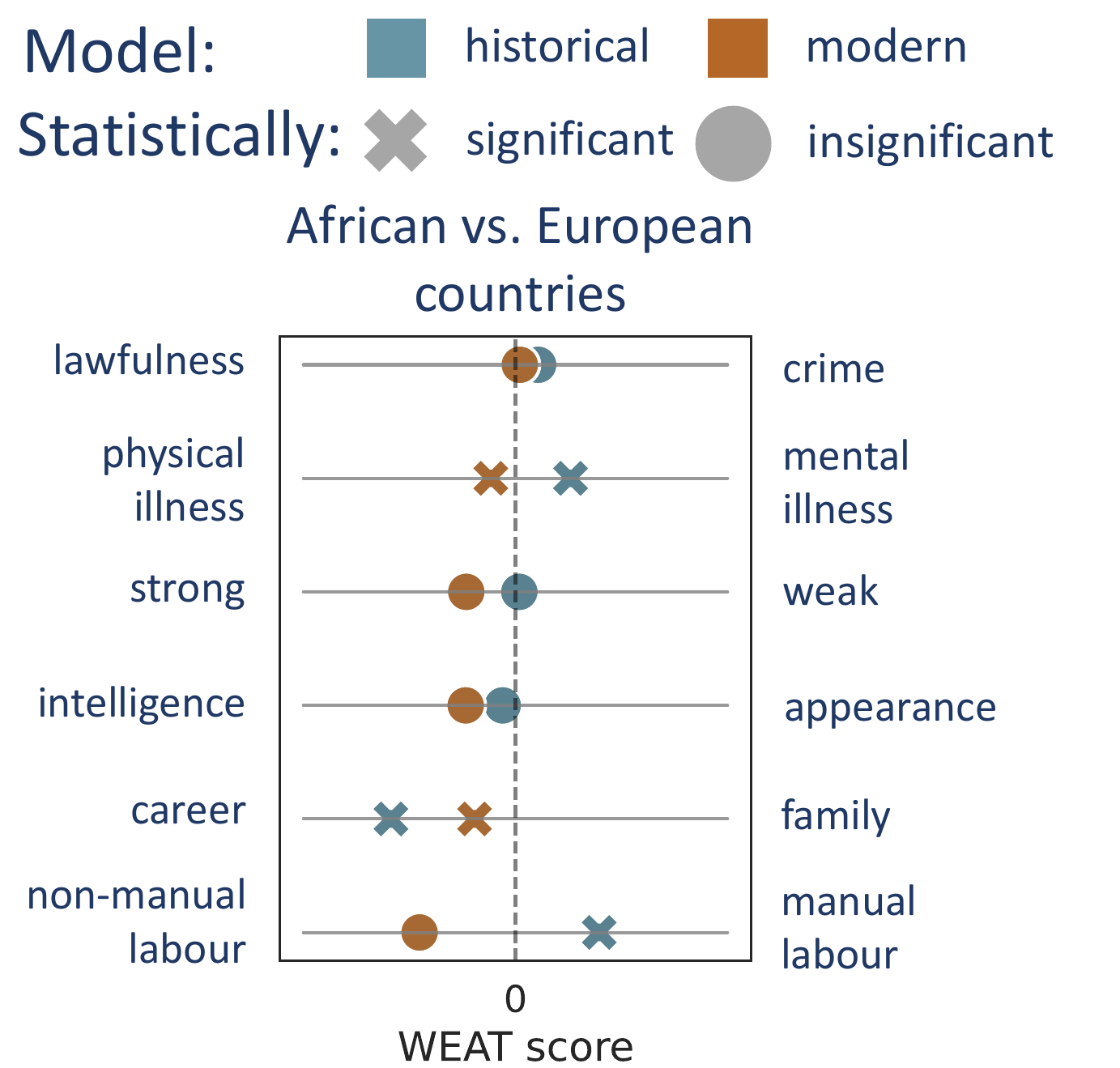}
         \caption{WEAT results of \textit{African countries} vs \textit{European countries}.}
         \label{fig:weat_3}
         
\end{figure}

\begin{figure}[ht]
    \centering

        \includegraphics[width=\columnwidth, trim={0cm 5cm 15.4cm 0cm},clip]{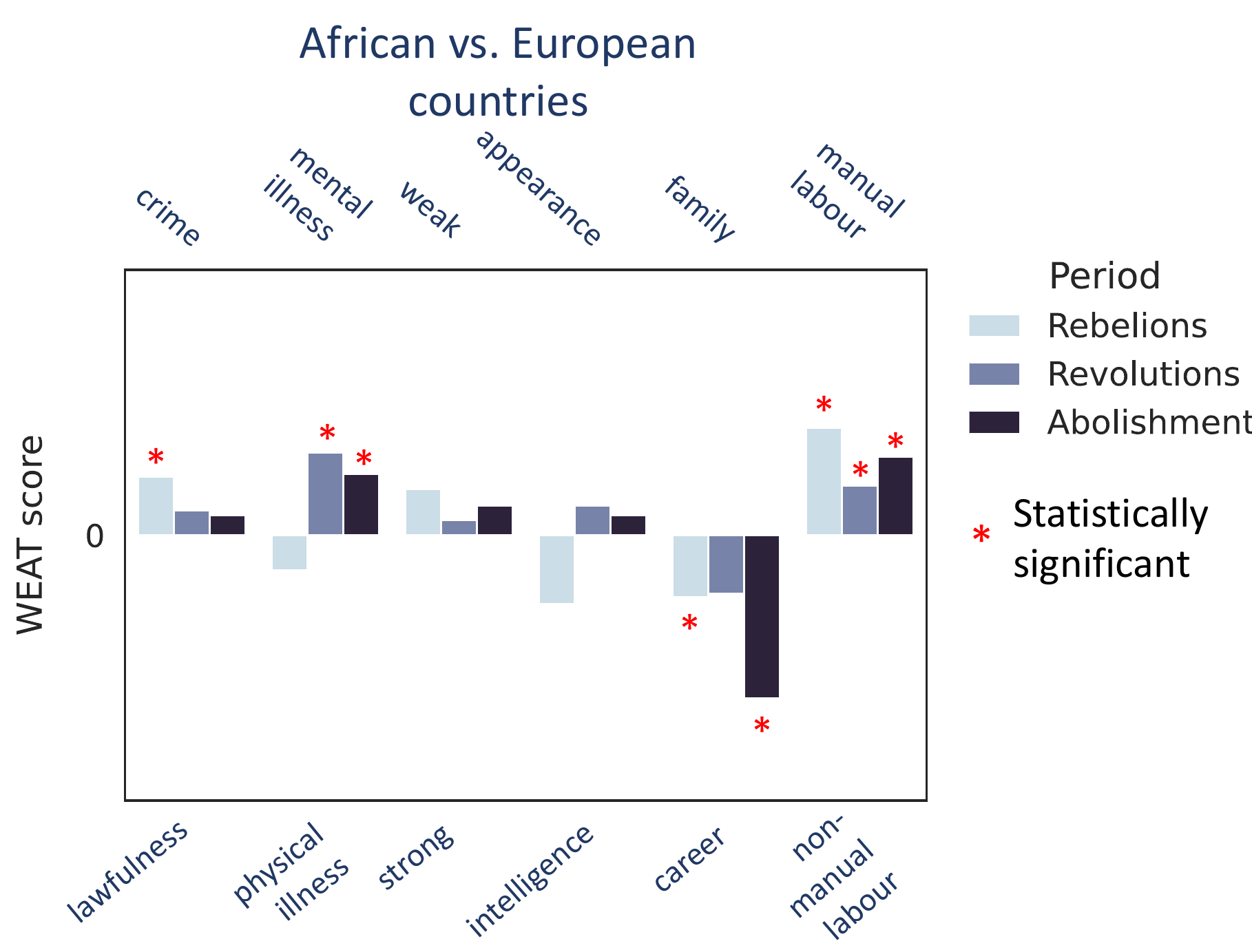}
         \caption{Temporal WEAT analysis conducted for the periods 1751--1790 (rebellions), 1791--1825 (revolutions) and 1826--1876 (abolishment). Similar to \Cref{fig:weat_all}, the height of each bar represents how strong the association of the attribute of \textit{African countries} is with each concept.}
         \label{fig:weat_temp_africa}
         
\end{figure}

In \Cref{tab:classification_acc}, we report the accuracy of the classified entities using the keyword-based approach. In \Cref{tab:interesting_examples}, we list examples of sentences from our newspaper dataset. \Cref{fig:weat_3} presents the WEAT results of the attributes \textit{African countries} vs \textit{European countries}. \Cref{fig:weat_temp_africa} presents temporal WEAT analysis conducted for the attributes \textit{African countries} vs \textit{European countries}. 

\begin{table*}[t]
\centering
\fontsize{10}{10}\selectfont
 \begin{tabular}{lrrr}
    \toprule
    Attribute & Ratio of correctly classified entities & Ratio of incorrectly classified entities & Ratio of unable to classify \\ \midrule
        Non-whites	& 0.89 & 0.036 & 0.07 \\
        Whites 	& 0.75 & 0.18 & 0.07 \\
        Males	& 0.89 & 0.036 & 0.07 \\
        Females	& 0.79 & 0.21 & 0 \\
    
    \bottomrule %
    \end{tabular}
     \caption{Performance of the keyword-based classification approach.}
     \label{tab:classification_acc}
\end{table*}

\begin{table*}[t]
\centering
\fontsize{10}{10}\selectfont
 \begin{tabular}{lp{10cm}}
    \toprule
    Word & Sentence \\ \midrule 
    
    ingenious & This comprehensive piece of clockwork cost the \textbf{ingenious} and indefatigable artist (one Jacob Lovelace, of Exeter,) 34 years’ labour. \\
    
    elderly & y un away for upwards of 16 Months past;; \textbf{elderly} NEGRO WOMAN hamed LOUISA, belongifg to the Estate of the late Ancup. \\
    
    active & FOR SALE, STRONG \textbf{active} NEGRO GIRL, about 24 Years of Age, she is a good Cook, can W asu, [rron, and is well acquainted with Housework in general. \\
    
    beautiful & and the young husband was hurried away, being scarcely permitted to take a parting kiss from his blooming and \textbf{beautiful} bride. \\
    
    blind & Dick, of the Mundingo Counrry, \textbf{blind} mark, about 18 years of ane, says he belongs te the estate Of ee Nichole, dec. of Mantego bay. \\

    sick & The young wife had snatched upa,; few of her own and her baby’s clothes; the husband, | Openiug Chorus, though \textbf{sick}, had attended to his duty to the last, and es | Song caped penniless with the clothes on his back. \\ 

    free & A \textbf{free} black girl JOSEPHINE, detained by the Police as being diseased; Proprietors and Managers an the Country are kindly requested to have the said Josephine apprehended ‘and lodged in the Towa Prison, the usual reward will be paid \\

    brave & From that moment the \textbf{brave} Lopez Lara was only occupied in devising means for delivering this notorious criminal into the hvids of justice. \\
    \bottomrule %

 \end{tabular}
 \caption{Examples from our dataset that contain biased words. Notice the high levels of noise and OCR errors.}
 \label{tab:interesting_examples}
\end{table*}

\end{document}